%% file: Ausarbeitung-Thoma.tex
\documentclass[technote,a4paper,leqno]{IEEEtran}
\pdfoutput=1
\usepackage[utf8]{inputenc} 
\usepackage[ngerman]{babel} 
\usepackage[T1]{fontenc}    
\usepackage{graphicx}
\usepackage{url}
\usepackage{amsmath}
\usepackage{amssymb}
\usepackage{enumerate}
\usepackage{scrtime}
\usepackage{listings}
\usepackage{hyperref}
\usepackage{cite}
\usepackage{parskip}
\usepackage[framed,amsmath,thmmarks,hyperref]{ntheorem}
\usepackage{algorithm}
\usepackage[noend]{algpseudocode}
\usepackage{csquotes}
\usepackage{subfig}         
\usepackage{caption}
\usepackage{tikz}

\usepackage{enumitem}
\usepackage[german]{cleveref} 
\usepackage{braket}
\allowdisplaybreaks
\usetikzlibrary{backgrounds}
\usepackage[binary-units=true]{siunitx}
\usepackage{mystyle}
\usepackage{microtype}

\hypersetup{
  pdftitle    = {Über die Klassifizierung von Knoten in dynamischen Netzwerken mit textuellen Inhalten},
  pdfauthor   = {Martin Thoma}, 
  pdfkeywords = {DYCOS}
}

\begin{document}

\title{Über die Klassifizierung von Knoten in dynamischen Netzwerken mit Inhalt}
\author{Martin Thoma}
\date{17.01.2014}
\maketitle

\begin{abstract}%
\input{abstract}
\end{abstract}

\section{Einleitung}
\input{Einleitung}

\section{Related Work}
\input{Related-Work}

\section{DYCOS}
\input{DYCOS-Algorithmus}

\section{Analyse des DYCOS-Algorithmus}
\input{Analyse}

\section{Probleme des DYCOS-Algorithmus}
\input{SchwaechenVerbesserungen}

\section{Ausblick}
\input{Ausblick}

\bibliographystyle{IEEEtranSA}
\bibliography{literatur}

\end{document}

%% file: abstract.tex
In dieser Arbeit wird der DYCOS-Algorithmus, wie er in \cite{aggarwal2011}
vorgestellt wurde, erklärt. Er arbeitet auf Graphen, deren Knoten teilweise mit
Beschriftungen versehen sind und ergänzt automatisch Beschriftungen für Knoten,
die bisher noch keine Beschriftung haben. Dieser Vorgang wird
\enquote{Klassifizierung} genannt. Dazu verwendet er die Struktur des Graphen
sowie textuelle Informationen, die den Knoten zugeordnet sind. Die in
\cite{aggarwal2011} beschriebene experimentelle Analyse ergab, dass er auch auf
dynamischen Graphen mit 19\,396 bzw. 806\,635 Knoten, von denen nur
14\,814 bzw. 18\,999 beschriftet waren, innerhalb von weniger als
einer Minute auf einem Kern einer Intel Xeon 2.5\,GHz~CPU mit 32\,G~RAM
ausgeführt werden kann.\\
Zusätzlich wird \cite{aggarwal2011} kritisch Erörtert und und es werden
mögliche Erweiterungen des DYCOS-Algorithmus vorgeschlagen.

\textbf{Keywords:} DYCOS, Label Propagation, Knotenklassifizierung

%% file: Einleitung.tex
Im Folgenden werden in \cref{sec:Motivation} einige Beispiele, in denen
der DYCOS-Algorithmus Anwendung finden könnte, dargelegt. In
\cref{sec:Problemstellung} wird die Problemstellung formal definiert
und in \cref{sec:Herausforderungen} wird auf besondere Herausforderungen der
Aufgabenstellung hingewiesen.

\subsection{Motivation}\label{sec:Motivation}
Teilweise beschriftete Graphen sind allgegenwärtig. Publikationsdatenbanken mit
Publikationen als Knoten, Literaturverweisen und Zitaten als Kanten sowie von
Nutzern vergebene Beschriftungen (sog. {\it Tags}) oder Kategorien als
Knotenbeschriftungen; Wikipedia mit Artikeln als Knoten, Links als Kanten und
Kategorien als Knotenbeschriftungen sowie soziale Netzwerke mit Eigenschaften
der Benutzer als Knotenbeschriftungen sind drei Beispiele dafür. Häufig sind
Knotenbeschriftungen nur teilweise vorhanden und es ist wünschenswert die
fehlenden Knotenbeschriftungen automatisiert zu ergänzen.

\subsection{Problemstellung}\label{sec:Problemstellung}
Gegeben ist ein Graph, dessen Knoten teilweise beschriftet sind. Zusätzlich
stehen zu einer Teilmenge der Knoten Texte bereit. Gesucht sind nun
Knotenbeschriftungen für alle Knoten, die bisher noch nicht beschriftet sind.\\

\begin{definition}[Knotenklassifierungsproblem]\label{def:Knotenklassifizierungsproblem}
    Sei $G_t = (V_t, E_t, V_{L,t})$ ein gerichteter Graph, wobei $V_t$ die
    Menge aller Knoten, $E_t \subseteq V_t \times V_t$ die Kantenmenge und
    $V_{L,t} \subseteq V_t$ die Menge der beschrifteten Knoten jeweils zum
    Zeitpunkt $t$ bezeichne. Außerdem sei $L_t$ die Menge aller zum Zeitpunkt
    $t$ vergebenen Knotenbeschriftungen und $f:V_{L,t} \rightarrow L_t$ die
    Funktion, die einen Knoten auf seine Beschriftung abbildet.

    Weiter sei für jeden Knoten $v \in V$ eine (eventuell leere) Textmenge
    $T(v)$ gegeben.

    Gesucht sind nun Beschriftungen für $V_t \setminus V_{L,t}$, also
    $\tilde{f}: V_t \setminus V_{L,t} \rightarrow L_t$. Die Aufgabe, zu $G_t$
    die Funktion $\tilde{f}$ zu finden heißt
    \textit{Knotenklassifierungsproblem}.
\end{definition}

\subsection{Herausforderungen}\label{sec:Herausforderungen}
Die Graphen, für die dieser Algorithmus konzipiert wurde, sind viele
$\num{10000}$~Knoten groß und dynamisch. \enquote{Dynamisch} bedeutet in diesem
Kontext, dass neue Knoten und eventuell auch neue Kanten hinzu kommen bzw.
Kanten oder Knoten werden entfernt werden. Außerdem stehen textuelle Inhalte zu
den Knoten bereit, die bei der Klassifikation genutzt werden können. Bei
kleinen Änderungen sollte nicht alles nochmals berechnen werden müssen, sondern
basierend auf zuvor berechneten Knotenbeschriftungen sollte die Klassifizierung
angepasst werden.

%% file: Related-Work.tex
Sowohl das Problem der Knotenklassifikation, als auch das der
Textklassifikation, wurde bereits in verschiedenen Kontexten analysiert. Jedoch
scheinen bisher entweder nur die Struktur des zugrundeliegenden Graphen oder
nur Eigenschaften der Texte verwendet worden zu sein.

So werden in \cite{bhagat,szummer} unter anderem Verfahren zur
Knotenklassifikation beschrieben, die wie der in \cite{aggarwal2011}
vorgestellte DYCOS-Algorithmus, um den es in dieser Ausarbeitung geht, auch auf
Random Walks basieren.

Obwohl es auch zur Textklassifikation einige Paper gibt
\cite{Zhu02learningfrom,Jiang2010302}, geht doch keines davon auf den
Spezialfall der Textklassifikation mit einem zugrundeliegenden Graphen ein.

Die vorgestellten Methoden zur Textklassifikation variieren außerdem sehr
stark. Es gibt Verfahren, die auf dem bag-of-words-Modell basieren
\cite{Ko:2012:STW:2348283.2348453} wie es auch im DYCOS-Algorithmus verwendet
wird. Aber es gibt auch Verfahren, die auf dem
Expectation-Maximization-Algorithmus basieren \cite{Nigam99textclassification}
oder Support Vector
Machines nutzen \cite{Joachims98textcategorization}.

Es wäre also gut Vorstellbar, die Art und Weise wie die Texte in die
Klassifikation des DYCOS-Algorithmus einfließen zu variieren. Allerdings ist
dabei darauf hinzuweisen, dass die im Folgenden vorgestellte Verwendung der
Texte sowohl einfach zu implementieren ist und nur lineare Vorverarbeitungszeit
in Anzahl der Wörter des Textes hat, als auch es erlaubt einzelne Knoten zu
klassifizieren, wobei der Graph nur lokal um den zu klassifizierenden Knoten
betrachten werden muss.

%% file: DYCOS-Algorithmus.tex
\subsection{Überblick}
DYCOS (\underline{DY}namic \underline{C}lassification algorithm with
c\underline{O}ntent and \underline{S}tructure) ist ein
Knotenklassifizierungsalgorithmus, der Ursprünglich in \cite{aggarwal2011}
vorgestellt wurde.

Ein zentrales Element des DYCOS-Algorithmus ist der sog. {\it Random Walk}:

\begin{definition}[Random Walk, Sprung]
    Sei $G = (V, E)$ mit $E \subseteq V \times V$ ein Graph und
    $v_0 \in V$ ein Knoten des Graphen.

    Ein Random Walk der Länge $l$ auf $G$, startend bei $v_0$ ist nun der
    zeitdiskrete stochastische Prozess, der $v_i$ auf einen zufällig gewählten
    Nachbarn $v_{i+1}$ abbildet (für $i \in 0, \dots, l-1$). Die Abbildung $v_i
    \mapsto v_{i+1}$ heißt ein Sprung.
\end{definition}

Der DYCOS-Algorithmus klassifiziert einzelne Knoten, indem $r$ Random Walks der
Länge $l$, startend bei dem zu klassifizierenden Knoten $v$ gemacht werden.
Dabei werden die Beschriftungen der besuchten Knoten gezählt. Die Beschriftung,
die am häufigsten vorgekommen ist, wird als Beschriftung für $v$ gewählt. DYCOS
nutzt also die sog. Homophilie, d.~h. die Eigenschaft, dass Knoten, die nur
wenige Hops von einander entfernt sind, häufig auch ähnlich sind \cite{bhagat}.
Der DYCOS-Algorithmus arbeitet jedoch nicht direkt auf dem Graphen, sondern
erweitert ihn mit Hilfe der zur Verfügung stehenden Texte. Wie diese
Erweiterung erstellt wird, wird im Folgenden erklärt.\\
Für diese Erweiterung wird zuerst wird Vokabular $W_t$ bestimmt, das
charakteristisch für eine Knotengruppe ist. Wie das gemacht werden kann und
warum nicht einfach jedes Wort in das Vokabular aufgenommen wird, wird in
\cref{sec:vokabularbestimmung} erläutert.\\
Nach der Bestimmung des Vokabulars wird für jedes Wort im Vokabular ein
Wortknoten zum Graphen hinzugefügt. Alle Knoten, die der Graph zuvor hatte,
werden nun \enquote{Strukturknoten} genannt.
Ein Strukturknoten $v$ wird genau dann mit einem Wortknoten $w \in W_t$
verbunden, wenn $w$ in einem Text von $v$ vorkommt. \Cref{fig:erweiterter-graph}
zeigt beispielhaft den so entstehenden, semi-bipartiten Graphen.
Der DYCOS-Algorithmus betrachtet also die Texte, die einem Knoten
zugeordnet sind, als eine Multimenge von Wörtern. Das heißt, zum einen
wird nicht auf die Reihenfolge der Wörter geachtet, zum anderen wird
bei Texten eines Knotens nicht zwischen verschiedenen
Texten unterschieden. Jedoch wird die Anzahl der Vorkommen
jedes Wortes berücksichtigt.

\begin{figure}[htp]
    \centering
    \input{figures/graph-content-and-structure.tex}
    \caption{Erweiterter Graph}
    \label{fig:erweiterter-graph}
\end{figure}

Entsprechend werden zwei unterschiedliche Sprungtypen unterschieden, die
strukturellen Sprünge und inhaltliche Zweifachsprünge:

\begin{definition}[struktureller Sprung]
    Sei $G_{E,t} = (V_t, E_{S,t} \cup E_{W,t}, V_{L,t}, W_{t})$ der
    um die Wortknoten $W_{t}$ erweiterte Graph.

    Dann heißt das zufällige wechseln des aktuell betrachteten
    Knoten $v \in V_t$ zu einem benachbartem Knoten $w \in V_t$
    ein \textit{struktureller Sprung}.
\end{definition}
\goodbreak
Im Gegensatz dazu benutzten inhaltliche Zweifachsprünge tatsächlich die
Grapherweiterung:
\begin{definition}[inhaltlicher Zweifachsprung]
    Sei $G_t = (V_t, E_{S,t} \cup E_{W,t}, V_{L,t}, W_{t})$ der um die
    Wortknoten $W_{t}$ erweiterte Graph.

    Dann heißt das zufällige wechseln des aktuell betrachteten Knoten $v \in
    V_t$ zu einem benachbartem Knoten $w \in W_t$ und weiter zu einem
    zufälligem Nachbar $v' \in V_t$ von $w$ ein inhaltlicher Zweifachsprung.
\end{definition}

Jeder inhaltliche Zweifachsprung beginnt und endet also in einem
Strukturknoten, springt über einen Wortknoten und ist ein Pfad der Länge~2.

Ob in einem Sprung der Random Walks ein struktureller Sprung oder ein
inhaltlicher Zweifachsprung gemacht wird, wird jedes mal zufällig neu
entschieden. Dafür wird der Parameter $0 \leq p_S \leq 1$ für den Algorithmus
gewählt. Mit einer Wahrscheinlichkeit von $p_S$ wird ein struktureller Sprung
durchgeführt und mit einer Wahrscheinlichkeit von $(1-p_S)$ ein modifizierter
inhaltlicher Zweifachsprung, wie er in \cref{sec:sprungtypen} erklärt wird,
gemacht. Der Parameter $p_S$ gibt an, wie wichtig die Struktur des Graphen im
Verhältnis zu den textuellen Inhalten ist. Bei $p_S = 0$ werden ausschließlich
die Texte betrachtet, bei $p_S = 1$ ausschließlich die Struktur des Graphen.

Die Vokabularbestimmung kann zu jedem Zeitpunkt $t$ durchgeführt werden, muss
es aber nicht.

In \cref{alg:DYCOS} steht der DYCOS-Algorithmus in Form von Pseudocode:
In \cref{alg1:l8} wird für jeden unbeschrifteten Knoten
durch die folgenden Zeilen eine Beschriftung gewählt.

\Cref{alg1:l10} führt $r$ Random Walks durch. In \cref{alg1:l11} wird eine
temporäre Variable für den aktuell betrachteten Knoten angelegt.

In \cref{alg1:l12} bis \cref{alg1:l21} werden einzelne Random Walks der Länge
$l$ durchgeführt, wobei die beobachteten Beschriftungen gezählt werden und mit
einer Wahrscheinlichkeit von $p_S$ ein struktureller Sprung durchgeführt wird.

\begin{algorithm}[ht]
    \begin{algorithmic}[1]
        \Require \\$G_{E,t} = (V_t, E_{S,t} \cup E_{W,t}, V_{L,t}, W_t)$ (Erweiterter Graph),\\
                 $r$ (Anzahl der Random Walks),\\
                 $l$ (Länge eines Random Walks),\\
                 $p_s$ (Wahrscheinlichkeit eines strukturellen Sprungs),\\
                 $q$ (Anzahl der betrachteten Knoten in der Clusteranalyse)
        \Ensure  Klassifikation von $V_t \setminus V_{L,t}$\\
        \\

        \ForAll{Knoten $v \in V_t \setminus V_{L,t}$}\label{alg1:l8}
            \State $d \gets $ leeres assoziatives Array
            \For{$i = 1, \dots,r$}\label{alg1:l10}
                \State $w \gets v$\label{alg1:l11}
                \For{$j= 1, \dots, l$}\label{alg1:l12}
                    \State $sprungTyp \gets \Call{random}{0, 1}$
                    \If{$sprungTyp \leq p_S$}
                        \State $w \gets$ \Call{SturkturellerSprung}{$w$}
                    \Else
                        \State $w \gets$ \Call{InhaltlicherZweifachsprung}{$w$}
                    \EndIf
                    \State $beschriftung \gets w.\Call{GetLabel}{ }$
                    \If{$!d.\Call{hasKey}{beschriftung}$}
                        \State $d[beschriftung] \gets 0$
                    \EndIf
                    \State $d[beschriftung] \gets d[beschriftung] + 1$
                \EndFor\label{alg1:l21}
            \EndFor

            \If{$d$.\Call{isEmpty}{ }} \Comment{Es wurde kein beschrifteter Knoten gesehen}
                \State $M_H \gets \Call{HäufigsteLabelImGraph}{ }$
            \Else
                \State $M_H \gets \Call{max}{d}$
            \EndIf
            \\
            \State \Comment{Wähle aus der Menge der häufigsten Beschriftungen $M_H$ zufällig eine aus}
            \State $label \gets \Call{Random}{M_H}$
            \State $v.\Call{AddLabel}{label}$ \Comment{und weise dieses $v$ zu}
        \EndFor
        \State \Return Beschriftungen für $V_t \setminus V_{L,t}$
    \end{algorithmic}
\caption{DYCOS-Algorithmus}
\label{alg:DYCOS}
\end{algorithm}

\subsection{Datenstrukturen}
Zusätzlich zu dem gerichteten Graphen $G_t = (V_t, E_t, V_{L,t})$ verwaltet der
DYCOS-Algorithmus zwei weitere Datenstrukturen:
\begin{itemize}
    \item Für jeden Knoten $v \in V_t$ werden die vorkommenden Wörter,
          die auch im Vokabular $W_t$ sind,
          und deren Anzahl gespeichert. Das könnte z.~B. über ein
          assoziatives Array (auch \enquote{dictionary} oder
            \enquote{map} genannt) geschehen. Wörter, die nicht in
          Texten von $v$ vorkommen, sind nicht im Array. Für
          alle vorkommenden Wörter ist der gespeicherte Wert zum
          Schlüssel $w \in W_t$ die Anzahl der Vorkommen von
          $w$ in den Texten von $v$.
    \item Für jedes Wort des Vokabulars $W_t$ wird eine Liste von
          Knoten verwaltet, in deren Texten das Wort vorkommt.
          Diese Liste wird bei den inhaltlichen Zweifachsprung,
          der in \cref{sec:sprungtypen} erklärt wird,
          verwendet.
\end{itemize}

\input{Sprungtypen}
\input{Vokabularbestimmung}

%% file: figures/graph-content-and-structure.tex
\tikzstyle{vertex}=[draw,black,circle,minimum size=10pt,inner sep=0pt]
\tikzstyle{edge}=[very thick]
\begin{tikzpicture}[scale=1.3]
    \node (a)[vertex] at (0,0) {};
    \node (b)[vertex]  at (0,1) {};
    \node (c)[vertex] at (0,2) {};
    \node (d)[vertex] at (1,0) {};
    \node (e)[vertex]  at (1,1) {};
    \node (f)[vertex] at (1,2) {};
    \node (g)[vertex] at (2,0) {};
    \node (h)[vertex] at (2,1) {};
    \node (i)[vertex] at (2,2) {};

    \node (x)[vertex] at (4,0) {};
    \node (y)[vertex] at (4,1) {};
    \node (z)[vertex] at (4,2) {};

    \draw[edge] (a) -- (d);
    \draw[edge] (b) -- (d);
    \draw[edge] (b) -- (c);
    \draw[edge] (c) -- (d);
    \draw[edge] (d) -- (e);
    \draw[edge] (d) edge[bend left] (f);
    \draw[edge] (d) edge[bend right] (x);
    \draw[edge] (g) edge (x);
    \draw[edge] (h) edge (x);
    \draw[edge] (h) edge (y);
    \draw[edge] (h) edge (e);
    \draw[edge] (e) edge (z);
    \draw[edge] (i) edge (y);

    \draw [dashed] (-0.3,-0.3) rectangle (2.3,2.3);
    \draw [dashed] (2.5,2.3) rectangle (5, -0.3);

    \node (struktur)[label={[label distance=0cm]0:Sturkturknoten $V_t$}] at (-0.1,2.5) {};
    \node (struktur)[label={[label distance=0cm]0:Wortknoten $W_t$}] at (2.7,2.5) {};
\end{tikzpicture}

%% file: Sprungtypen.tex
\subsection{Sprungtypen}\label{sec:sprungtypen}
Die beiden bereits definierten Sprungtypen, der strukturelle Sprung sowie der
inhaltliche Zweifachsprung werden im Folgenden erklärt.
\goodbreak
Der strukturelle Sprung entspricht einer zufälligen Wahl eines Nachbarknotens,
wie es in \cref{alg:DYCOS-structural-hop} gezeigt wird.
\begin{algorithm}[H]
    \begin{algorithmic}[1]
        \Procedure{SturkturellerSprung}{Knoten $v$, Anzahl $q$}
            \State $n \gets v.\Call{NeighborCount}{}$ \Comment{Wähle aus der Liste der Nachbarknoten}
            \State $r \gets \Call{RandomInt}{0, n-1}$ \Comment{einen zufällig aus}
            \State $v \gets v.\Call{Next}{r}$ \Comment{Gehe zu diesem Knoten}
            \State \Return $v$
        \EndProcedure
    \end{algorithmic}
\caption{Struktureller Sprung}
\label{alg:DYCOS-structural-hop}
\end{algorithm}

Bei inhaltlichen Zweifachsprüngen ist jedoch nicht sinnvoll so strikt nach der
Definition vorzugehen,  also direkt von einem strukturellem Knoten $v \in V_t$
zu einem mit $v$ verbundenen Wortknoten $w \in W_t$ zu springen und von diesem
wieder zu einem verbundenem strukturellem Knoten $v' \in V_t$. Würde dies
gemacht werden, wäre zu befürchten, dass aufgrund von Homonymen die Qualität der
Klassifizierung verringert wird. So hat \enquote{Brücke} im Deutschen viele
Bedeutungen. Gemeint sein können z.~B. das Bauwerk, das Entwurfsmuster der
objektorientierten Programmierung oder ein Teil des Gehirns.

Deshalb wird für jeden Knoten $v$, von dem aus ein inhaltlicher
Zweifachsprung gemacht werden soll folgende Textanalyse durchgeführt:
\begin{enumerate}[label=C\arabic*,ref=C\arabic*]
    \item \label{step:c1} Gehe alle in $v$ startenden Random Walks der Länge $2$ durch
          und erstelle eine Liste $L$ der erreichbaren Knoten $v'$. Speichere
          außerdem, durch wie viele Pfade diese Knoten $v'$ jeweils erreichbar
          sind.
    \item \label{step:c2} Betrachte im Folgenden nur die Top-$q$ Knoten bzgl.
          der Anzahl der Pfade von $v$ nach $v'$, wobei $q \in \mathbb{N}$
          eine zu wählende Konstante des DYCOS-Algorithmus ist.\footnote{Sowohl für den DBLP, als auch für den
CORA-Datensatz wurde in \cite[S. 364]{aggarwal2011} $q=10$ gewählt.}
          Diese Knotenmenge heiße im Folgenden $T(v)$ und $p(v, v')$ sei die
          Anzahl der Pfade von $v$ über einen Wortknoten nach $v'$.
    \item \label{step:c3} Wähle mit Wahrscheinlichkeit
          $\frac{p(v, v')}{\sum_{w \in T(v)} p(v, w)}$ den Knoten $v' \in T(v)$
          als Ziel des Zweifachsprungs.
\end{enumerate}

Konkret könnte also ein inhaltlicher Zweifachsprung sowie wie in
\cref{alg:DYCOS-content-multihop} beschrieben umgesetzt werden.
Der Algorithmus bekommt einen Startknoten $v \in V_T$ und
einen $q \in \mathbb{N}$ als Parameter. $q$ ist ein Parameter der
für den DYCOS-Algorithmus zu wählen ist. Dieser Parameter beschränkt
die Anzahl der möglichen Zielknoten $v' \in V_T$ auf diejenigen
$q$ Knoten, die $v$ bzgl. der Textanalyse am ähnlichsten sind.

In \cref{alg:l2} bis \cref{alg:l5} wird \cref{step:c1} durchgeführt und alle
erreichbaren Knoten in $reachableNodes$ mit der Anzahl der Pfade, durch die sie
erreicht werden können, gespeichert.

In \cref{alg:l6} wird \cref{step:c2} durchgeführt. Ab hier gilt
\[ |T| = \begin{cases}q               &\text{falls } |reachableNodes|\geq q\\
                     |reachableNodes| &\text{sonst }\end{cases}\]

Bei der Wahl der Datenstruktur von $T$ ist zu beachten, dass man in
\cref{alg:21} über Indizes auf Elemente aus $T$ zugreifen können muss.

In \cref{alg:l8} bis \ref{alg:l13} wird ein assoziatives Array erstellt, das
von $v' \in T(v)$ auf die relative Häufigkeit bzgl. aller Pfade von $v$ zu
Knoten aus den Top-$q$ abbildet.

In allen folgenden Zeilen wird \cref{step:c3} durchgeführt. In \cref{alg:15}
bis \cref{alg:22} wird ein Knoten $v' \in T(v)$ mit einer Wahrscheinlichkeit,
die seiner relativen Häufigkeit am Anteil der Pfaden der Länge 2 von $v$ nach
$v'$ über einen beliebigen Wortknoten entspricht ausgewählt und schließlich
zurückgegeben.

\begin{algorithm}
  \caption{Inhaltlicher Zweifachsprung}
  \label{alg:DYCOS-content-multihop}
    \begin{algorithmic}[1]
        \Procedure{InhaltlicherZweifachsprung}{Knoten $v \in V_T$, $q \in \mathbb{N}$}
            \State $erreichbareKnoten \gets$ leeres assoziatives Array\label{alg:l2}
            \ForAll{Wortknoten $w$ in $v.\Call{getWordNodes}{ }$}
                \ForAll{Strukturknoten $x$ in $w.\Call{getStructuralNodes}{ }$}
                    \If{$!erreichbareKnoten.\Call{hasKey}{x}$}
                        \State $erreichbareKnoten[x] \gets 0$
                    \EndIf
                    \State $erreichbareKnoten[x] \gets erreichbareKnoten[x] + 1$
                \EndFor
            \EndFor\label{alg:l5}
            \State \label{alg:l6} $T \gets \Call{max}{erreichbareKnoten, q}$
            \\
            \State \label{alg:l8} $s \gets 0$
            \ForAll{Knoten $x \in T$}
                \State $s \gets s + erreichbareKnoten[x]$
            \EndFor
            \State $relativeHaeufigkeit \gets $ leeres assoziatives Array
            \ForAll{Knoten $x \in T$}
                \State $relativeHaeufigkeit \gets \frac{erreichbareKnoten[x]}{s}$
            \EndFor\label{alg:l13}
            \\
            \State \label{alg:15} $random \gets \Call{random}{0, 1}$
            \State $r \gets 0.0$
            \State $i \gets 0$
            \While{$s < random$}
                \State $r \gets r + relativeHaeufigkeit[i]$
                \State $i \gets i + 1$
            \EndWhile

            \State $v \gets T[i-1]$ \label{alg:21}
            \State \Return $v$ \label{alg:22}
        \EndProcedure
    \end{algorithmic}
\end{algorithm}

%% file: Vokabularbestimmung.tex
\subsection{Vokabularbestimmung}\label{sec:vokabularbestimmung}
Da die Größe des Vokabulars die Datenmenge signifikant beeinflusst,
liegt es in unserem Interesse so wenig Wörter wie möglich ins
Vokabular aufzunehmen. Insbesondere sind Wörter nicht von Interesse,
die in fast allen Texten vorkommen, wie im Deutschen z.~B.
\enquote{und}, \enquote{mit} und die Pronomen. Es ist wünschenswert Wörter zu
wählen, die die Texte möglichst stark voneinander Unterscheiden. Der
DYCOS-Algorithmus wählt die Top-$m$ dieser Wörter als Vokabular, wobei
$m \in \mathbb{N}$ eine festzulegende Konstante ist. In \cite[S. 365]{aggarwal2011}
wird der Einfluss von $m \in \Set{5,10, 15,20}$ auf die Klassifikationsgüte
untersucht und festgestellt, dass die Klassifikationsgüte mit größerem $m$
sinkt, sie also für $m=5$ für den DBLP-Datensatz am höchsten ist. Für den
CORA-Datensatz wurde mit $m \in \set{3,4,5,6}$ getestet und kein signifikanter
Unterschied festgestellt.

Nun kann man manuell eine Liste von zu beachtenden Wörtern erstellen
oder mit Hilfe des Gini-Koeffizienten automatisch ein Vokabular erstellen.
Der Gini-Koeffizient ist ein statistisches Maß, das die Ungleichverteilung
bewertet. Er ist immer im Intervall $[0,1]$, wobei $0$ einer
Gleichverteilung entspricht und $1$ der größtmöglichen Ungleichverteilung.

Sei nun $n_i(w)$ die Häufigkeit des Wortes $w$ in allen Texten mit der $i$-ten
Knotenbeschriftung.
\begin{align}
    p_i(w) &:= \frac{n_i(w)}{\sum_{j=1}^{|\L_t|} n_j(w)} &\text{(Relative Häufigkeit des Wortes $w$)}\\
    G(w)   &:= \sum_{j=1}^{|\L_t|} p_j(w)^2              &\text{(Gini-Koeffizient von $w$)}
\end{align}
In diesem Fall ist $G(w)=0$ nicht möglich, da zur Vokabularbestimmung nur
Wörter betrachtet werden, die auch vorkommen.

Ein Vorschlag, wie die Vokabularbestimmung implementiert werden kann, ist als
Pseudocode mit \cref{alg:vokabularbestimmung} gegeben. In \cref{alg4:l6} wird
eine Teilmenge $S_t \subseteq V_{L,t}$ zum Generieren des Vokabulars gewählt.
In \cref{alg4:l8} wird ein Array $cLabelWords$ erstellt, das $(|\L_t|+1)$
Felder hat. Die Elemente dieser Felder sind jeweils assoziative Arrays, deren
Schlüssel Wörter und deren Werte natürliche Zahlen sind. Die ersten $|\L_t|$
Elemente von $cLabelWords$ dienen dem Zählen der Häufigkeit der Wörter von
Texten aus $S_t$, wobei für jede Beschriftung die Häufigkeit einzeln gezählt
wird. Das letzte Element aus $cLabelWords$ zählt die Summe der Wörter. Diese
Datenstruktur wird in \cref{alg4:l10} bis \ref{alg4:l12} gefüllt.

In \cref{alg4:l17} bis \ref{alg4:l19} wird die relative Häufigkeit der Wörter
bzgl. der Beschriftungen bestimmt. Daraus wird in \cref{alg4:l20} bis
\ref{alg4:l22} der Gini-Koeffizient berechnet. Schließlich werden in
\cref{alg4:l23} bis \ref{alg4:l24} die Top-$q$ Wörter mit den
höchsten Gini-Koeffizienten zurückgegeben.

\begin{algorithm}[ht]
    \begin{algorithmic}[1]
        \Require \\
                 $V_{L,t}$ (beschriftete Knoten),\\
                 $\L_t$ (Menge der Beschriftungen),\\
                 $f:V_{L,t} \rightarrow \L_t$ (Beschriftungsfunktion),\\
                 $m$ (Gewünschte Vokabulargröße)
        \Ensure  $\M_t$ (Vokabular)\\
        \State $S_t \gets \Call{Sample}{V_{L,t}}$\label{alg4:l6} \Comment{Wähle $S_t \subseteq V_{L,t}$ aus}
        \State $\M_t \gets \emptyset$ \Comment{Menge aller Wörter}
        \State $cLabelWords \gets$ Array aus $(|\L_t|+1)$ assoziativen Arrays\label{alg4:l8}
        \ForAll{$v \in S_t$} \label{alg4:l10}
            \State $i \gets \Call{getLabel}{v}$
            \State \Comment{$w$ ist das Wort, $c$ ist die Häufigkeit}
            \ForAll{$(w, c) \in \Call{getTextAsMultiset}{v}$}
                \State $cLabelWords[i][w] \gets cLabelWords[i][w] + c$
                \State $cLabelWords[|\L_t|][w] \gets cLabelWords[i][|\L_t|] + c$
                \State $\M_t = \M_t \cup \Set{w}$
            \EndFor
        \EndFor\label{alg4:l12}
		\\
        \ForAll{Wort $w \in \M_t$}
            \State $p \gets $ Array aus $|\L_t|$ Zahlen in $[0, 1]$\label{alg4:l17}
            \ForAll{Label $i \in \L_t$}
                \State $p[i] \gets \frac{cLabelWords[i][w]}{cLabelWords[i][|\L_t|]}$
            \EndFor\label{alg4:l19}

            \State $w$.gini $\gets 0$ \label{alg4:l20}
            \ForAll{$i \in 1, \dots, |\L_t|$}
                \State $w$.gini $\gets$ $w$.gini + $p[i]^2$
            \EndFor\label{alg4:l22}
        \EndFor

        \State $\M_t \gets \Call{SortDescendingByGini}{\M_t}$\label{alg4:l23}
        \State \Return $\Call{Top}{\M_t, m}$\label{alg4:l24}
    \end{algorithmic}
\caption{Vokabularbestimmung}
\label{alg:vokabularbestimmung}
\end{algorithm}

Die Menge $S_t$ kann aus der Menge aller Dokumente, deren Knoten beschriftet
sind, mithilfe des in \cite{Vitter} vorgestellten Algorithmus bestimmt werden.

%% file: Analyse.tex
Für den DYCOS-Algorithmus wurde in \cite{aggarwal2011} bewiesen, dass sich nach
Ausführung von DYCOS für einen unbeschrifteten Knoten mit einer
Wahrscheinlichkeit von höchstens $(|\L_t|-1)\cdot e^{-l \cdot b^2 / 2}$ eine
Knotenbeschriftung ergibt, deren relative Häufigkeit weniger als $b$ der
häufigsten Beschriftung ist. Dabei ist $|\L_t|$ die Anzahl der Beschriftungen
und $l$ die Länge der Random-Walks.

Außerdem wurde experimentell anhand des DBLP-Datensatzes\footnote{http://dblp.uni-trier.de/}
und des CORA-Datensatzes\footnote{http://people.cs.umass.edu/~mccallum/data/cora-classify.tar.gz}
gezeigt (vgl. \cref{tab:datasets}), dass die Klassifikationsgüte nicht wesentlich von der Anzahl der Wörter mit
höchstem Gini-Koeffizient $m$ abhängt. Des Weiteren betrug die Ausführungszeit
auf einem Kern eines Intel Xeon $\SI{2.5}{\GHz}$ Servers mit
$\SI{32}{\giga\byte}$~RAM für den DBLP-Datensatz unter $\SI{25}{\second}$,
für den CORA-Datensatz sogar unter $\SI{5}{\second}$. Dabei wurde eine
für CORA eine Klassifikationsgüte von \SIrange{82}{84}{\percent} und
auf den DBLP-Daten von \SIrange{61}{66}{\percent} erreicht.

\begin{table}[htp]
    \centering
    \begin{tabular}{|l||r|r|r|r|}\hline
    \textbf{Name} & \textbf{Knoten} & \textbf{davon beschriftet} & \textbf{Kanten}  & \textbf{Beschriftungen} \\ \hline\hline
    \textbf{CORA} & \num{19396}  & \num{14814}             & \num{75021}   & 5              \\
    \textbf{DBLP} & \num{806635} & \num{18999 }            & \num{4414135} & 5              \\\hline
    \end{tabular}
    \caption{Datensätze, die für die experimentelle Analyse benutzt wurden}
    \label{tab:datasets}
\end{table}

Obwohl es sich nicht sagen lässt, wie genau die Ergebnisse aus
\cite{aggarwal2011} zustande gekommen sind, eignet sich das
Kreuzvalidierungsverfahren zur Bestimmung der Klassifikationsgüte wie es in
\cite{Lavesson,Stone1974} vorgestellt wird:
\begin{enumerate}
    \item Betrachte nur $V_{L,T}$.
    \item Unterteile $V_{L,T}$ zufällig in $k$ disjunkte Mengen $M_1, \dots, M_k$.
    \item \label{schritt3} Teste die Klassifikationsgüte, wenn die Knotenbeschriftungen
          aller Knoten in $M_i$ für DYCOS verborgen werden für $i=1,\dots, k$.
    \item Bilde den Durchschnitt der Klassifikationsgüten aus \cref{schritt3}.
\end{enumerate}

Es wird $k=10$ vorgeschlagen.

%% file: SchwaechenVerbesserungen.tex
Bei der Anwendung des in \cite{aggarwal2011} vorgestellten Algorithmus auf
reale Datensätze könnten zwei Probleme auftreten, die im Folgenden erläutert
werden. Außerdem werden Verbesserungen vorgeschlagen, die es allerdings noch zu
untersuchen gilt.

\subsection{Anzahl der Knotenbeschriftungen}
So, wie der DYCOS-Algorithmus vorgestellt wurde, können nur Graphen bearbeitet
werden, deren Knoten jeweils höchstens eine Beschriftung haben. In vielen
Fällen, wie z.~B. Wikipedia mit Kategorien als Knotenbeschriftungen haben
Knoten jedoch viele Beschriftungen.

Auf einen ersten Blick ist diese Schwäche einfach zu beheben, indem man beim
zählen der Knotenbeschriftungen für jeden Knoten jedes Beschriftung zählt. Dann
wäre noch die Frage zu klären, mit wie vielen Beschriftungen der betrachtete
Knoten beschriftet werden soll.

Jedoch ist z.~B. bei Wikipedia-Artikeln auf den Knoten eine Hierarchie
definiert. So ist die Kategorie \enquote{Klassifikationsverfahren} eine
Unterkategorie von \enquote{Klassifikation}. Bei dem Kategorisieren von
Artikeln sind möglichst spezifische Kategorien vorzuziehen, also kann man nicht
einfach bei dem Auftreten der Kategorie \enquote{Klassifikationsverfahren}
sowohl für diese Kategorie als auch für die Kategorie \enquote{Klassifikation}
zählen.

\subsection{Überanpassung und Reklassifizierung}
\begin{figure}[ht]
    \centering
    \subfloat[$t=1$]{
        \input{figures/graph-t1.tex}
        \label{fig:graph-t1}
    }%
    \subfloat[$t=2$]{
        \input{figures/graph-t2.tex}
        \label{fig:graph-t2}
    }

    \subfloat[$t=3$]{
        \input{figures/graph-t3.tex}
        \label{fig:graph-t3}
    }%
    \subfloat[$t=4$]{
        \input{figures/graph-t4.tex}
        \label{fig:graph-t4}
    }%
    \caption{Minimalbeispiel für den Einfluss früherer DYCOS-Anwendungen}
    \label{fig:Formen}
\end{figure}
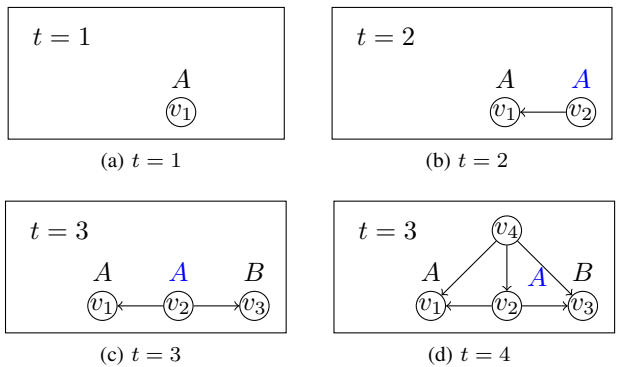

Aggarwal und Li beschreiben in \cite{aggarwal2011} nicht, auf welche Knoten der
Klassifizierungsalgorithmus angewendet werden soll. Jedoch ist die Reihenfolge
der Klassifizierung relevant. Dazu folgendes Minimalbeispiel:

Gegeben sei ein dynamischer Graph ohne textuelle Inhalte. Zum Zeitpunkt $t=1$
habe dieser Graph genau einen Knoten $v_1$ und $v_1$  sei mit dem $A$
beschriftet. Zum Zeitpunkt $t=2$ komme ein nicht beschrifteter Knoten $v_2$
sowie die Kante $(v_2, v_1)$ hinzu.\\
Nun wird der DYCOS-Algorithmus auf diesen Knoten angewendet und $v_2$ mit $A$
beschriftet.\\
Zum Zeitpunkt $t=3$ komme ein Knoten $v_3$, der mit $B$ beschriftet ist, und
die Kante $(v_2, v_3)$ hinzu. \Cref{fig:Formen} visualisiert diesen Vorgang.

Würde man nun den DYCOS-Algorithmus erst jetzt, also anstelle von Zeitpunkt
$t=2$ zum Zeitpunkt $t=3$ auf den Knoten $v_2$ anwenden, so würde eine
\SI{50}{\percent}-Wahrscheinlichkeit bestehen, dass dieser mit $B$ beschriftet
wird. Aber in diesem Beispiel wurde der Knoten schon zum Zeitpunkt $t=2$
beschriftet. Obwohl es in diesem kleinem Beispiel noch keine Rolle spielt, wird
das Problem klar, wenn man weitere Knoten einfügt:

Wird zum Zeitpunkt $t=4$ ein unbeschrifteter Knoten $v_4$ und die Kanten
$(v_1, v_4)$, $(v_2, v_4)$, $(v_3, v_4)$ hinzugefügt, so ist die
Wahrscheinlichkeit, dass $v_4$ mit $A$ beschriftet wird bei $\frac{2}{3}$.
Werden die unbeschrifteten Knoten jedoch erst jetzt und alle gemeinsam
beschriftet, so ist die Wahrscheinlichkeit für $A$ als Beschriftung bei nur $50\%$.
Bei dem DYCOS-Algorithmus findet also eine Überanpassung an vergangene
Beschriftungen statt.

Das Reklassifizieren von Knoten könnte eine mögliche Lösung für dieses
Problem sein. Knoten, die durch den DYCOS-Algorithmus beschriftet wurden
könnten eine Lebenszeit bekommen (TTL, Time to Live). Ist diese
abgelaufen, wird der DYCOS-Algorithmus erneut auf den Knoten angewendet.

%% file: figures/graph-t1.tex
\tikzstyle{vertex}=[draw,black,circle,minimum size=10pt,inner sep=0pt]
\tikzstyle{edge}=[very thick]
\begin{tikzpicture}[scale=1,framed]
    \node (a)[vertex,label=$A$] at (0,0) {$v_1$};
    \node (b)[vertex, white] at (1,0) {$v_2$};
    \node (struktur)[label={[label distance=-0.2cm]0:$t=1$}] at (-2,1) {};
\end{tikzpicture}

%% file: figures/graph-t2.tex
\tikzstyle{vertex}=[draw,black,circle,minimum size=10pt,inner sep=0pt]
\tikzstyle{edge}=[very thick]
\begin{tikzpicture}[scale=1,framed]
    \node (a)[vertex,label=$A$] at (0,0) {$v_1$};
    \node (b)[vertex,label={\color{blue}$A$}] at (1,0) {$v_2$};
    \draw[->] (b) -- (a);
    \node (struktur)[label={[label distance=-0.2cm]0:$t=2$}] at (-2,1) {};
\end{tikzpicture}

%% file: figures/graph-t3.tex
\tikzstyle{vertex}=[draw,black,circle,minimum size=10pt,inner sep=0pt]
\tikzstyle{edge}=[very thick]
\begin{tikzpicture}[scale=1,framed]
    \node (a)[vertex,label=$A$] at (0,0) {$v_1$};
    \node (b)[vertex,label={\color{blue}$A$}] at (1,0) {$v_2$};
    \node (c)[vertex,label=$B$] at (2,0) {$v_3$};
    \draw[->] (b) -- (a);
    \draw[->] (b) -- (c);
    \node (struktur)[label={[label distance=-0.2cm]0:$t=3$}] at (-1,1) {};
\end{tikzpicture}

%% file: figures/graph-t4.tex
\tikzstyle{vertex}=[draw,black,circle,minimum size=10pt,inner sep=0pt]
\tikzstyle{edge}=[very thick]
\begin{tikzpicture}[scale=1,framed]
    \node (a)[vertex,label=$A$] at (0,0) {$v_1$};
    \node (b)[vertex,label=45:{\color{blue}$A$}] at (1,0) {$v_2$};
    \node (c)[vertex,label=$B$] at (2,0) {$v_3$};
    \node (d)[vertex] at (1,1) {$v_4$};
    \draw[->] (b) -- (a);
    \draw[->] (b) -- (c);

    \draw[->] (d) -- (a);
    \draw[->] (d) -- (b);
    \draw[->] (d) -- (c);
    \node (struktur)[label={[label distance=-0.2cm]0:$t=3$}] at (-1,1) {};
\end{tikzpicture}

%% file: Ausblick.tex
Den DYCOS-Algorithmus kann in einigen Aspekten erweitert werden. So könnte man
vor der Auswahl des Vokabulars jedes Wort auf den Wortstamm zurückführen. Dafür
könnte zum Beispiel der in \cite{porter} vorgestellte Porter-Stemming-Algorithmus verwendet werden. Durch diese Maßnahme wird das Vokabular kleiner
gehalten wodurch mehr Artikel mit einander durch Vokabular verbunden werden
können. Außerdem könnte so der Gini-Koeffizient ein besseres Maß für die
Gleichheit von Texten werden.

Eine weitere Verbesserungsmöglichkeit besteht in der Textanalyse. Momentan ist
diese noch sehr einfach gestrickt und ignoriert die Reihenfolge von Wörtern
beziehungsweise Wertungen davon. So könnte man den DYCOS-Algorithmus in einem
sozialem Netzwerk verwenden wollen, in dem politische Parteiaffinität von
einigen Mitgliedern angegeben wird um die Parteiaffinität der restlichen
Mitglieder zu bestimmen. In diesem Fall macht es jedoch einen wichtigen
Unterschied, ob jemand über eine Partei gutes oder schlechtes schreibt.

Eine einfache Erweiterung des DYCOS-Algorithmus wäre der Umgang mit mehreren
Beschriftungen.

DYCOS beschränkt sich bei inhaltlichen Zweifachsprüngen auf die
Top-$q$-Wortknoten, also die $q$ ähnlichsten Knoten gemessen mit der
Aggregatanalyse, allerdings wurde bisher noch nicht untersucht, wie der
Einfluss von $q \in \mathbb{N}$ auf die Klassifikationsgüte ist.